\documentclass{article}

\usepackage{threeparttable}
\usepackage{arxiv}
\usepackage[utf8]{inputenc} 
\usepackage[T1]{fontenc}    
\usepackage{hyperref}       
\usepackage{url}            
\usepackage{booktabs}       
\usepackage{amsfonts}       
\usepackage{nicefrac}       
\usepackage{microtype}      
\usepackage{lipsum}
\usepackage{graphicx}

\usepackage{cite}

\title{An Enhanced Prohibited Items Recognition Model}

\author{
 Tianze Rong, Hongxiang Cai, Yichao Xiong \\
  \texttt{dominirong@gmail.com, chxlll@126.com, xyc\_sjtu@163.com}
}

\begin{document}
\maketitle

\section{Introduction}
\par Most of the security inspection contains prohibited items recognition. And the approach to solve the package checking is to scan the package or bag to acquire the ability of looking inside to recognize and locate the prohibited items. But it is high-maintenance and high-time-consuming to keep professional staff to read X-ray images and recognize the items in the images. Nowadays X-ray reading is rapid and automatic, benefiting from the development of deep neural networks and computer vision.
\par We investigated data of this field, SIXray as a typical dataset, to probe the characteristics of X-ray images and prohibited items recognition. 
\par In this work we found the primary factor that restricts the model performance is that the object scale is so small that reins the enhancing module. Then we investigated a bundle of data augmentation and enhancing modules to improve the performance of the model.
\par Our model can achieve a state of the art performance at SIXray——the best mAP of our model at SIXray10 is 0.899, at SIXray100 is 0.748.
\section{Method}
\subsection{Perspective of Data}
\subsubsection{Data Description}
\par The description and analysis to data can be representative of the data from similar tasks. In this case, we analyzed the statistics of SIXray dataset\cite{SIXray} and regard it as a typical pattern to research the bottleneck of prohibited items recognition via X-ray.

\paragraph{Amount of Images}
SIXray has three subsets in different imbalance levels called SIXray10, SIXray100 and SIXray1000. The imbalance level of SIXray1000 is almost as high as SIXray100. We only investigate the SIXray10 and SIXray100. Following Table \ref{tab:amount} is the size of the dataset and its subsets.
\begin{table}[htb]
    \centering
    \begin{tabular}{ccc}
    \hline
        &SIXray10 & SIXray100 \\
    \hline
         Train Set& 74959& 749574\\
         Test Set& 13411 & 133201\\
    \hline
        Total & 88370 & 882775\\
    \hline
    \\
    \end{tabular}
    \caption{Quantitative Statistics of SIXray}
    \label{tab:amount}
\end{table}
\par 

\paragraph{Statistics of Labels} Label imbalanced is one of dominant characteristics in prohibited items detection. It is intuitive to assume that prohibited items are much more rare than regular ones so that it is inevitable to estimate the level of imbalance. Table \ref{tab:label} shows the amount of each label.
\begin{table}[htb]
    \centering
    \begin{tabular}{ccccccc}
    \hline
        &Gun & Knife & Wrench & Pliers & Scissors & Negative\\
    \hline
         Counts & 2705& 1748 & 2012 & 3434 & 807 & 67464\\
         Percentage(\%) & 3.60 & 2.33 & 2.68 & 4.58 & 1.08 & 90.0\\

    \hline
    \\
    \end{tabular}
    \caption{Label Distribution of SIXray on Train Set}
    \label{tab:label}
\end{table}
\subsection{Other Features of Dataset}
\subsubsection{Small Object Recognition}
\begin{figure}[htb]
    \centering
    \includegraphics[scale=0.6]{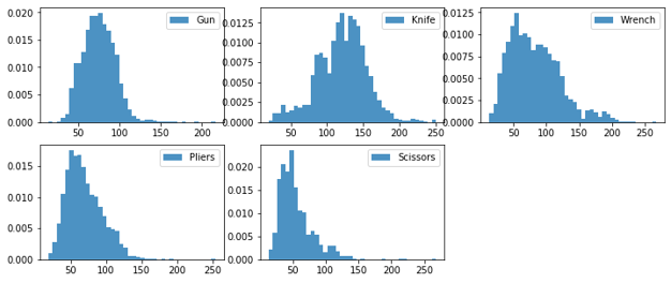}
    \caption{Scale Histograms of Categories}
    \label{fig:scale}
\end{figure}

Some of the prohibited items are fairly small-sized to be recognized so we investigated the dataset to estimate the scale of the object. Fortunately, the author of SIXray has built the detection annotations. For each of bounding box we calculated the scale by the formula:
$$scale = \sqrt{width*height}$$
Then we displayed the scales from different categories in their own histogram as in Figure \ref{fig:scale}
It is apparent that pliers and scissors are much smaller than the other categories. The most likely scale of scissors is about 50 pixels.  
\subsubsection{Overlapping}
According to \cite{SIXray}, X-ray images are transparent and even the item is occluded which can also be seen. The absorption of X-ray traveling through obeys Lambert-Beer's Law which points out that the  attenuation of light(X-ray included) in a transparent object is linear:
$$A=\epsilon lc$$
Where $A$ is the absorbance of light, $\epsilon$ is a coefficient related to attenuating species, $l$ is the traveling length of light, $c$ is the concentration of the attenuating species. Hence, it is natural to blend two images by linear overlay due to the transparency of X-ray images.
\begin{figure}[htb]
    \centering
    \includegraphics[scale=0.6]{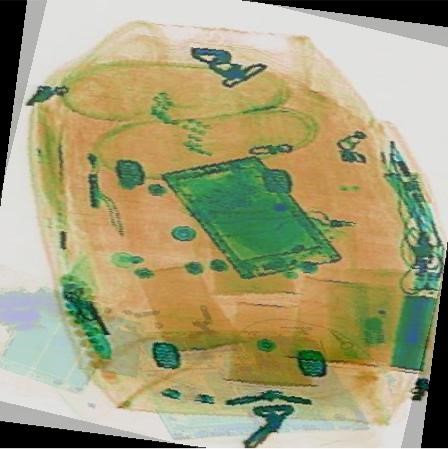}
    \caption{Data Augmentation about Overlapping}
    \label{fig:mixup}
\end{figure}
\subsection{Solution to SIXray}
\subsubsection{Data Augmentation}
\paragraph{Random Flipping}Since the X-ray image is perspective and taken from the vertical view which makes the pose of packages can be variant and arbitrary. At least, the vertical flipping and horizontal flipping will not harm the semantics.
\paragraph{Random Rotation} The same reason with random flipping but the random rotation can be more variant than random flipping.
\paragraph{Random Cropping} The X-ray image mostly has a blank margin around the X-ray, meanwhile the object to be recognized is not placed right in the middle of the image. As we draw the random crop into the data augmentation, whether the margin or the position of the object can be a prerequisite. In addition, we sampled some image to check the possibility, which is seldom occur, of cropping the object out.
\paragraph{Image Synthesis} Since the data is imbalanced, oversampling is a good way to ease the imbalance, what's more, adding two samples with weight can be an option benefiting from the transparency of X-ray image. Therefore, we use the weighted adding as an augmentation, which is called blending by us. You can see in Figure \ref{fig:mixup}. The mixup is also an option to enlarge the dataset capacity:
$$Image_{Blend}=Blend(Image_1, Image_2, \lambda)=\lambda Image_1+(1-\lambda)Image_2$$
Here $\lambda$ is a super-parameter.
\subsubsection{Imbalanced Label}
We designed a rescoring mechanism against the imbalance between classes\cite{YOLO9000}. As we showed in Table \ref{tab:label}, the amount of negative sample is more than any of the amount of a single category of prohibited items, which means as we decouple the imbalance between positive and negative samples and the imbalance between positive classes can somehow ease the imbalance among the whole dataset. It can be regarded as a rather weak hierarchy structure.
\par To represent the imbalance between positive sample and negative sample, we adopted the probability of positive sample as the objectness score.
\par And based on the objectness score, we regress probability of the corresponding class by the formula: $$P(class_i)=P(class_i|object)*P(object)$$
\subsubsection{Attention Mechanism}
Prohibited items are mostly local and partial from the whole image. In a way, the majority of the image is uninformative or should be ignored. Attention mechanisms, especially spatial attention, can lead the model to focus on the local region to promote the performance of the model.
\par Spatial attention\cite{spatial} is widely used as a plug and play module to promote the performance of models which distribute a weight to each of the pixel height-wise and weight-wise. Prohibited items should be weighted more in this case. Channel-wise attention\cite{senet} is another form of attention on feature-channel. The Convolutional Block Attention Module(CBAM) is combined with the channel-wise and spatial attention.
\begin{figure}[htb]
    \centering
    \includegraphics[scale=0.3]{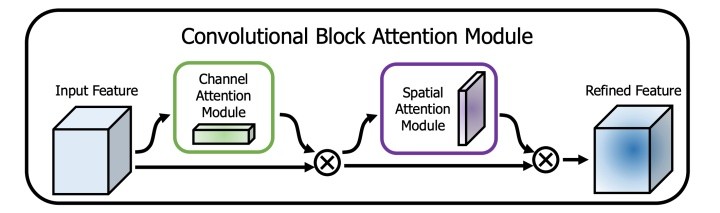}
    \caption{CBAM Structure\cite{CBAM}}
    \label{fig:my_label}
\end{figure}
\section{Experiments}
\subsection{SIXray10}
\subsubsection{Baseline}
\paragraph{Backbone} An ordinary ResNet-34\cite{resnet} architecture pre-trained on ImageNet\cite{imagenet}, but with sigmoid function to output the confidence as multi-label.
\paragraph{Loss Function} Intuitively we adopted the binary cross entropy loss as the loss function since this task is multi-label classification. 
\paragraph{Optimizer} The optimizer is Nesterov Accelerated Gradient(NAG), learning rate is set to 0.01 without learning rate scheduler. Also the momentum parameter is set to 0.9.
\paragraph{Data Augmentation} The input data will be random flipping on both vertical and horizontal direction with a probability of 0.5. Besides, all images will be resized to (224, 224).
\paragraph{Training Procedure} The batch size is 128 and training for 60 epochs. The training duration is optimal and selected as the best setting. The metrics of our baseline is shown below in Figure \ref{tab:baseline}.
\begin{table}[htb]
    \centering
    \begin{tabular}{ccccccc}
    \hline
        AP(\%) &Gun & Knife & Wrench & Pliers & Scissors & mean\\
    \hline

         ResNet34\cite{SIXray} & \textbf{89.7} & 85.5 & 62.8 & 83.5 & 53.0 & 74.8\\
         DenseNet-CHR\cite{SIXray} &  87.1 & 85.9 & 70.5 & 88.3 &66.1& 79.6 \\
         Baseline(Ours) & 89.5& \textbf{91.7} & \textbf{76.4} & \textbf{88.5} & \textbf{66.5} & \textbf{82.3}\\
    \hline
    \\
    \end{tabular}
    \caption{Metrics on Baseline Setting Trained on SIXray10}
    \label{tab:baseline}
\end{table}
\subsubsection{Data Augmentation}
\paragraph{Random Crop} Instead of directly resize to (224, 224), we firstly resize the image to a size of (256, 256), then randomly cropped to (224, 224).
\paragraph{Random Rotate} To rotate the image with a random degree between $(-15^{\circ}, 15^{\circ})$, and resize the image to keep all pixels are still in the region of image.
\paragraph{MixUp} MixUp\cite{mixup} is to mix two group images up with a partition coefficient $\lambda$ which is submitting to a beta distribution $B(\alpha, \beta)$. We add two different image via
adding with:
$$Image^i_{blending}=\lambda Image^i_{original}+(1-\lambda)Image^i_{shuffled}$$
Their loss is calculated by the formula:
$$Loss = \lambda Loss_{original}+(1-\lambda)Loss_{shuffled}$$
\paragraph{Blending} Due to the overlapping property, directly add two images with a constant coefficient $\lambda$, here$Image^i_{shuffled}$ means images from a certain batch after shuffling:
$$Image^i_{blending}=\lambda Image^i_{original}+(1-\lambda)Image^i_{shuffled}$$
corresponding label is:
$$label_{blending}= label_{original} | label_{shuffled}$$
where | is dimension-wise-or.

\begin{table}[htb]
    \centering
    \begin{tabular}{cccccccc}
    \hline
    Method & Baseline &Random Crop & MixUp(0.2, 0.2)& MixUp(0.2, 0.2)& Blend(0.5)\\
    \hline
    mAP(\%) &82.3& 82.5&81.9&82.2&66.0\\

    \hline
    \\
    \end{tabular}
    \caption{Metrics on Different Data Augmentation}
    \label{tab:my_label}
\end{table}
\subsubsection{Attention Mechanism}
\begin{figure}[htb]
    \centering
    \includegraphics[scale=0.5]{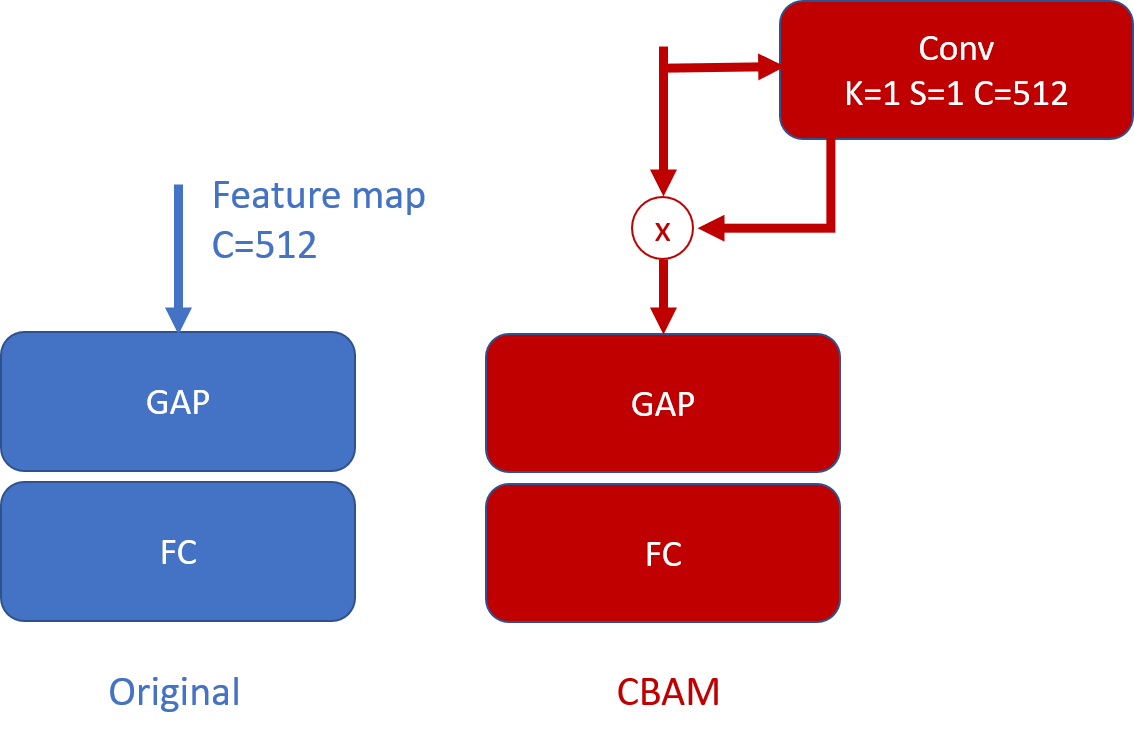}
    \caption{Implementation of CBAM Structure}
    \label{fig:CBAM}
\end{figure}

As stated before, the prohibited item could be aimed by attention mechanism. We employed the Convolutional Block Attention Module(CBAM)\cite{CBAM} as the attention mechanism into our model. CBAM is an attention module with both channel-wise and pixel-wise attention. But we modified the implement from the original one, the structure is like Figure \ref{fig:CBAM} following.
\begin{table}[htb]
    \centering
    \begin{tabular}{ccc}
    \hline
         & Baseline & CBAM\\
    \hline
         mAP(\%) & 82.3 & 83.0 \\
    \hline
    \\
    \end{tabular}
    \caption{Result of CBAM}
    \label{tab:cbam}
\end{table}

\subsubsection{Input Scale}
\begin{figure}[htb]
    \centering
    \includegraphics[scale=0.8]{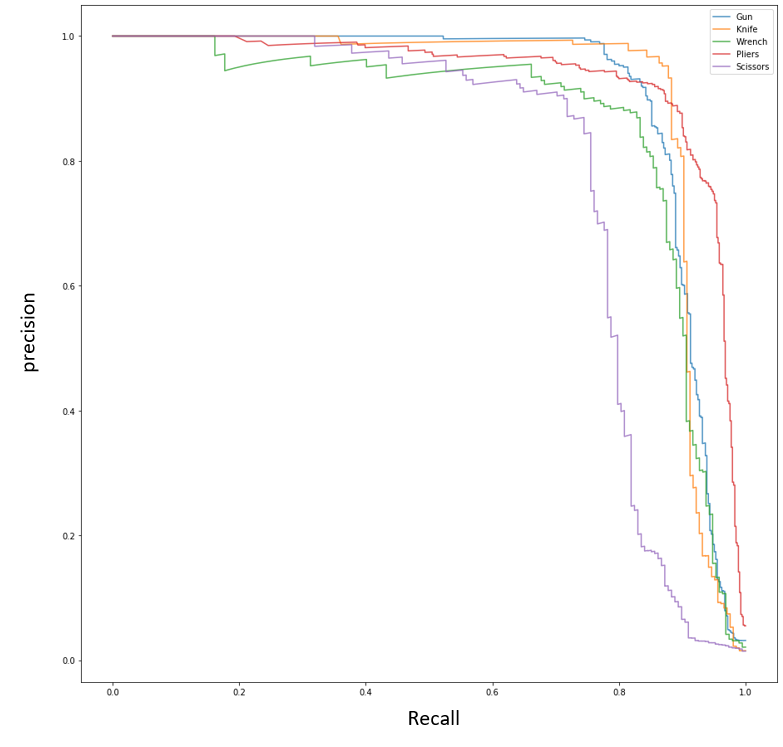}
    \caption{P-R curve of Baseline}
    \label{fig:pr}
\end{figure}

After the experiment we mentioned, we found it is counter-intuitive that the normal and universal methods to promote our model are all disabled. We checked most of the potential factors to check out the bottleneck. It can be clearly delivered from the baseline Table \ref{tab:baseline} and P-R curve in Figure \ref{fig:pr} that the category of scissors has a lower performance than the others and similarly the label of pliers has the second lower performance, particularly the recall of scissors is flopping on the curve. 
\par To associate with the statistics of objects, the label of scissors is exactly the smallest category while the pliers is second. We change the input scale into (384, 384) and (512, 512) and enlarge the crop size with equal ratio.
\begin{table}[htb]
    \centering
    \begin{tabular}{ccc}
    \hline
         Input Size & Crop Size & mAP(\%)\\
    \hline
         256 & 224 & 82.3 \\
         384 & 336 & 85.8 \\
         512 & 448 & \textbf{87.8} \\ 
    \hline
    \\
    \end{tabular}
    \caption{Result under Different Input Scale}
    \label{tab:my_label}
\end{table}

\subsubsection{Summary}
After the scale-related experiment, we found the scale of images maybe a performance bottleneck holds the accuracy off. We ran a set of experiments to release the restriction and revalidate those methods that did not work before. Here is some description of notes might be used:
\begin{itemize}
    \item Input scale: The edge length of images after resizing.
    \item Crop scale: The  edge length of images after random cropping.
    \item Flip: Using the random flip on vertical and horizontal direction with probability of 0.5.     
    \item Rotation: Using the random rotation with a rotating angle between $(-15^{\circ}, 15^{\circ})$.
    \item Synthesis: Using the image synthesized with MixUp or blending.
    \item CBAM: Using CBAM module on classification head.

\end{itemize}
\begin{table}[htb]
    \centering
    {
    \begin{tabular}{ccccccc}
    \hline
         Input Scale & Crop Scale & Flip & Rotation & Synthesis & CBAM & mAP\\
    \hline
        224 & &\checkmark &&&&82.3\\
        256 & 224 &\checkmark & && &\textbf{82.5} \\
        256 & 224 &\checkmark & && \checkmark&\textbf{83.0} \\
        256 & 224 &\checkmark & \checkmark &&&81.8 \\
        256 & 224 &\checkmark & &MixUp(0.2, 0.2) &&81.9\\
        256 & 224 &\checkmark & &MixUp(0.4, 0.4) &&82.2\\
        256 & 224 &\checkmark & & Blend(0.5) &&66.0\\
        384 & 336 &\checkmark &&&&\textbf{85.8}\\
        512 & 448 &\checkmark &&&&\textbf{87.8}\\
        512 & 448 &\checkmark & \checkmark &&&\textbf{88.7} \\
        512 & 448 &\checkmark &&&\checkmark&\textbf{88.0}\\
        512 & 448 &\checkmark & &MixUp(0.4, 0.4) &&\textbf{89.9}\\
        512 & 448 &\checkmark & & Blend(0.5) &&86.5\\
        512 & 448 &\checkmark &\checkmark &MixUp(0.4, 0.4) &&86.5\\
        512 & 448 &\checkmark & &MixUp(0.4, 0.4) &\checkmark&\textbf{89.9}\\
    \hline
    \\
    \end{tabular}}
    \caption{Results and Conditions of All Experiment}
    \label{tab:my_label}
\end{table}
Finally, we considered that the final setting is:
\begin{itemize}
    \item Scale: Input scale is (512, 512).
    \item Random Crop: Cropping scale is (448, 448).
    \item Random Flip: Using the random flip on vertical and horizontal direction with probability of 0.5.     
    \item CBAM Module on classification head.
    \item Mixup: Alpha and beta is 0.4.
\end{itemize}
\subsection{SIXray100}
\subsubsection{Baseline}
We tested the best model trained on SIXray10 and a model whose setting is inherited from the best SIXray10 but trained on SIXray 100.
\begin{table}[htb]
    \centering
    \begin{tabular}{ccccccc}
    \hline
        AP(\%) &Gun & Knife & Wrench & Pliers & Scissors & mean\\
    \hline
        ResNet34\cite{SIXray}  & 83.1 & 78.8 & 30.5 &55.2 &16.1& 52.7 \\
        DenseNet-CHR\cite{SIXray} &82.1&78.8 &43.2&66.8&28.8& 60.0\\
        Trained on SIXray100 & 82.0& {85.8} & \textbf{64.4} & \textbf{77.1} & 53.4 &72.6\\

        Trained on SIXray10 &\textbf{85.1}&\textbf{86.8}&61.4 &\textbf{77.1} &\textbf{57.0} & \textbf{73.5}\\
    \hline
    \\
    \end{tabular}
    \caption{Metrics on Baseline Setting Trained on SIXray100}
    \label{tab:my_label}
\end{table}
\subsubsection{Rescoring}
By analyzing the results of baseline on SIXray100, the set trained on SIXray100 is even worse than SIXray10 against the common sense that the bigger the data is, the better the model works. Furthermore the variant is controlled it is reasonable to believe that the model degradation is due to the higher imbalance level. The solution we designed is the rescoring mechanism.
\par We modified the output layer of the FC-layer to adapt the rescoring mechanism.
\par According to the label of SIXray the dimension of the output layer should be 5, equal to the number of categories. We modified the dimension into 6. The surplus is the objectness, which is to predict the probability if there is a prohibited item in the image but without classification. The consequent probability of each categories is the five components multiply with the objectness as the formula we mentioned before:
$$P(class_i)=P(class_i|object)*P(object)$$
\begin{table}[htb]
    \centering
    \begin{tabular}{ccccccc}
    \hline
        AP(\%) &Gun & Knife & Wrench & Pliers & Scissors & mean\\
    \hline

        Trained on SIXray10 &{85.1}&\textbf{86.8}&61.4 &{77.1} &{57.0} & {73.5}\\
        rescoring &\textbf{ 87.4}& {86.6} & \textbf{61.6} & \textbf{80.1} & \textbf{58.4} &\textbf{74.8}\\
    \hline
    \\
    \end{tabular}
    \caption{Metrics of Rescoring Mechanism}
    \label{tab:my_label}
\end{table}
\section{Conclusion}
Conclusively, we adopt a input scale of 512, crop scale of 448, and with random flip, CBAM and mix up whose alpha and beta is 0.4 as basic configuration can achieve a mAP of 89.9\% on SIXray10. As for SIXray100 need rescoring mechanism additionally, which achieve a  mAP of 74.8\% on SIXray100.
\bibliographystyle{unsrt}
\bibliography{template}

\end{document}